\documentclass[10pt, a4paper]{article}
\usepackage{lrec}
\usepackage{multibib}
\newcites{languageresource}{Language Resources}
\usepackage{graphicx}
\usepackage{tabularx}
\usepackage{slashbox}
\usepackage{soul}
\usepackage{multirow}

\usepackage{epstopdf}
\usepackage[latin1]{inputenc}

\usepackage{hyperref}
\usepackage{xstring}
\usepackage{eqnarray,amsmath}

\title{Upping the Ante: \\
Towards a Better Benchmark for Chinese-to-English Machine Translation}

\name{Christian Hadiwinoto, Hwee Tou Ng}

\address{Department of Computer Science \\
National University of Singapore \\
\{chrhad, nght\}@comp.nus.edu.sg\\}

\abstract{
There are many machine translation (MT) papers that propose novel approaches and show improvements over their self-defined baselines. The experimental setting in each paper often differs from one another. As such, it is hard to determine if a proposed approach is really useful and advances the state of the art. Chinese-to-English translation is a common translation direction in MT papers, although there is not one widely accepted experimental setting in Chinese-to-English MT. Our goal in this paper is to propose a benchmark in evaluation setup for Chinese-to-English machine translation, such that the effectiveness of a new proposed MT approach can be directly compared to previous approaches. Towards this end, we also built a highly competitive state-of-the-art MT system trained on a large-scale training set. Our system outperforms reported results on NIST OpenMT test sets in almost all papers published in major conferences and journals in computational linguistics and artificial intelligence in the past 11 years. We argue that a standardized benchmark on data and performance is important for meaningful comparison. \\ \newline  }

\begin{document}

\maketitleabstract

\section{Introduction}
\label{sec:introduction}

Over the years, there have been many published papers on machine translation (MT), proposing novel ideas by showing improvements over certain baselines. However, a paper often has a different experimental setup from the others. These variations include the approach (algorithm) and dataset. Ideally, research and development work on MT should be based on a benchmark of system setup with good performance. Otherwise, there is no support in asserting that a proposed method advances the state of the art.

Unfortunately, for Chinese-to-English MT,  two widely spoken human languages and one of the most well-studied language translation directions in MT, there is no widely accepted standard benchmark for evaluation, comprising a standardized training set, development set, and test set. Throughout the past decade, Chinese-to-English translation has been most commonly performed on NIST OpenMT\footnote{\url{www.nist.gov/itl/iad/mig/open-machine-translation-evaluation}} test sets, trained on parallel and monolingual corpora from the Linguistic Data Consortium (LDC)\footnote{\url{catalog.ldc.upenn.edu}}.

Our goal in this paper is to propose a benchmark in evaluation setup for Chinese-to-English machine translation, such that the effectiveness of a new proposed MT approach can be directly compared to previous approaches. Towards this end, we also built a highly competitive state-of-the-art MT system trained on a large-scale training set. Our system outperforms reported results on NIST OpenMT test sets in almost all papers published in major conferences and journals in computational linguistics and artificial intelligence in the past 11 years.

The rest of this paper is organized as follows. Section 2 describes our MT approach. Section 3 elaborates our experimental setup. Section 4 presents our experimental results. Section 5 describes related work. Finally, Section 6 gives the conclusion.

\section{Neural Machine Translation}
\label{sec:nmt}

We built a neural machine translation (NMT) system based on the encoder-decoder approach with attention mechanism \cite{bahdanau_neural_2015}. This NMT approach encodes an input sentence into a continuous representation by an encoder recurrent neural network (RNN) and produces translation output by a decoder RNN. The decoder RNN, through an attention mechanism, looks into different parts of the encoded input sentence while decoding is in progress.

\subsection{Encoder-Decoder Model with Attention}
\label{sec:nmt_basic}

Given a target language sentence ${\bf y}=(y_1,...,y_n)$ and the corresponding source language sentence ${\bf x}=(x_1,...,x_m)$, the neural machine translation model is formulated as
\begin{equation}
p({\bf y}|{\bf x}) = \prod_{i=1}^{n} p(y_i|y_1,...,y_{i-1},{\bf x})
\label{eq:nmtm_outseq}
\end{equation}
in which the probability of the target word $y_i$ at time step $i$ is computed by the decoder RNN as follows:
\begin{equation}
p(y_i|y_1,...,y_{i-1},{\bf x}) = F(y_i,y_{i-1},s_i,c_i) = t_i[y_i]
\label{eq:nmtm_outword}
\end{equation}
where $F$ is a function to compute the probability of the word $y_i$ to be generated at time step $i$ and $t_i$ is a vector having the size of the target language vocabulary, in which each vector dimension $t_i[y]$ stores the probability of a word $y$, computed as follows:
\begin{equation}
\label{eq:nmtm_outsoft}
t_i = \text{softmax}(W_t(\tanh(U_ts_i+V_tE[y_{i-1}]+C_tc_i+b_t)))
\end{equation}
where $U_t$, $V_t$, and $C_t$ are matrices mapping the hidden state $s_i$, the embedding of the previous word $E[y_{i-1}]$, and the context vector $c_i$ respectively to an intermediate vector representation, with $b_t$ being the bias vector. Then $W_t$ transforms the intermediate vector representation to a vocabulary-sized probability vector.

The decoder hidden state at a time step $i$ is computed by
\begin{equation}
s_i=g_y(E[y_{i-1}],s_{i-1},c_i)
\label{eq:nmtm_decode}
\end{equation}
where $g_y$ is the RNN unit function to compute the current hidden state given the hidden state of the previous time step, the previous word embedding, and the context.

Equation \ref{eq:nmtm_outsoft} indicates that the target word to be generated at a given time step takes into account the context vector $c_i$, which is a weighted sum of each annotation vector $h_j$, representing the source language sentence at position $j$:
\begin{equation}
\label{eq:nmtm_context}
c_i = \sum_{j=1}^{m}{\alpha_{ij}h_j}
\end{equation}
in which the scalar weight $\alpha_{ij}$ for each $h_j$ is computed by a softmax function:
\begin{equation}
\label{eq:nmtm_contextweight}
\alpha_{ij}=\frac{\exp(e_{ij})}{\sum_{k=1}^{m}{\exp(e_{ik})}}
\end{equation}
where $e_{ij}$ is computed by 
\begin{equation}
\label{eq:nmtm_matching}
e_{ij} = v_a^T \tanh(W_as'_i+U_ah_j+b_a) + \beta
\end{equation}
where $W_a$ and $U_a$ are the weight matrices and $b_a$ is the bias vector to compute a vector, which is then converted by the weight vector $v_a$ and the bias term $\beta$ into a scalar $e_{ij}$, i.e., the degree of matching between the target word at time step $i$ and the input word at position $j$. This is conceptually a soft alignment model.

To compute the decoding hidden state $s_i$ in Equation \ref{eq:nmtm_decode}, we adopt an approach that incorporates the context $c_i$ from the attention mechanism by using two transitions \cite{sennrich_nematus:_2017}. The decoder hidden state function $g_y(E[y_{i-1}],s_{i-1},c_i)$ in Equation \ref{eq:nmtm_decode} first passes the embedding $E[y_{i-1}]$ of the input word $y_{i-1}$ to the first recurrent unit function, resulting in an intermediate hidden state $s'_i$, which is computed by the decoder recurrent unit functions $g_{y,1}$ as:
\begin{equation}
s'_i = g_{y,1}(E[y_{i-1}], s_{i-1}) \label{eq:dec_basic}
\end{equation}
and is passed to Equation \ref{eq:nmtm_matching}. Then, the second recurrent unit function $g_{y,2}$ processes the context $c_i$ defined in Equation \ref{eq:nmtm_context} and the intermediate hidden state $s'_i$ as follows:
\begin{equation}
s_i = g_{y,2}(c_{i}, s'_{i}) = {g}_{att}(E[y_{i-1}],s_{i-1},c_i) \label{eq:dec_basic_att}
\end{equation}
where $g_{att}(E[y_{i-1}],s_{i-1},c_i)$ is a composition of $g_{y,1}$ and $g_{y,2}$. It is to be noted that as the decoder generates an output word $y_i$ at current time step $i$, the input word at the time step is $y_{i-1}$. The recurrent unit function is described further in Section \ref{sec:nmt_recunit}

We made use of the bidirectional encoder RNN, where each annotation vector $h_j$ is a concatenation of the forward and the backward RNN hidden states, $\overrightarrow{h_j}$ and $\overleftarrow{h_j}$ respectively, defined as follows:
\begin{align}
h_j &= [\overrightarrow{h}_{j};\overleftarrow{h}_{j}] \label{eq:nmtm_encode} \\
\overrightarrow{h}_{j} &= \overrightarrow{g}_{x}(E[x_j],\overrightarrow{h}_{j-1}) \label{eq:nmtm_encode_fw} \\
\overleftarrow{h}_{j} &= \overleftarrow{g}_{x}(E[x_j],\overleftarrow{h}_{j+1}) \label{eq:nmtm_encode_bw}
\end{align}
where $\overrightarrow{g}_{x}$ is the forward RNN unit function to compute the RNN hidden state at the current encoding position $j$ given the embedding of the current word $E[x_j]$ and the hidden state at the previous position, while $\overleftarrow{g}_{x}$ is the backward RNN unit function to compute the hidden state at $j$ given the word embedding $E[x_j]$ and the hidden state at the next position.

Training an end-to-end NMT model is conducted by the back-propagation through time (BPTT) algorithm, which updates the parameters of the RNN while the time steps are unrolled, to minimize a cost function. The parallel training corpus is divided into mini-batches, each consisting of $N$ parallel sentences. The NMT model parameters are updated in each mini-batch.

Translation decoding is performed by a beam search algorithm, which produces translation output sequentially in the target language order. NMT decoding proceeds by generating one word at each time step.

In NMT, as described in Equation \ref{eq:nmtm_outsoft}, computing the probability involves mapping the hidden state vector to a vector with the dimension of the vocabulary size. Therefore, to make computation tractable, the NMT vocabulary size is limited. To cope with the limitation of the vocabulary size, we adopt fragmentation of words into sub-words of character sequences through the byte pair encoding (BPE) algorithm \cite{sennrich_neural_2016}. This algorithm finds the $N$ most frequent character sequences of variable length, through $N$ character merge operations, and splits less frequent words based on this list of character sub-sequences.

\subsection{Recurrent Unit Function}
\label{sec:nmt_recunit}

To compute the hidden state representations in Equations \ref{eq:nmtm_decode} and \ref{eq:nmtm_encode}--\ref{eq:nmtm_encode_bw}, we made use of recurrent unit functions with gate mechanism to control the flow of information from the input and the previous hidden state. There are two commonly adopted gate mechanisms in the encoder-decoder RNN NMT model, namely the long short-term memory (LSTM) \cite{hochreiter_long_1997} and the gated recurrent unit (GRU) \cite{cho_learning_2014}.

The LSTM RNN unit consists of a memory cell $\mu_j$ and three gates, i.e., the input gate $\iota_j$ that controls the intensity of the new information to be stored in the memory cell, the forget gate $f_j$ that controls how much to remember or to forget from the previous memory cell, and the output gate $o_j$ that controls how much information is output to the hidden state from the memory. At each time step $j$, given the input $\chi_j$, the hidden state $\eta_j$ is formulated as:
\begin{align}
\eta_j &= {LSTM}(\chi_j,\eta_{j-1}) \nonumber \\
&= o_j \circ \tanh(\mu_j) \label{eq:lstm}
\end{align}
where
\begin{align*}
\iota_j &= \sigma(W_{\iota}\chi_j + U_{\iota}\eta_{j-1} + b_\iota) \\
f_j &= \sigma(W_{f}\chi_j + U_{f}{\eta}_{j-1} + b_f) \\
\mu_j &= f_j \circ \mu_{j-1} + \iota_j \circ \tanh(W_{\mu}\chi_j + U_{\mu}{\eta}_{j-1} + b_\mu) \\
o_j &= \sigma(W_{o}\chi_j + U_{o}{\eta}_{j-1} + b_o)
\end{align*}
$W$ and $U$ denote the weight matrices transforming the input embedding and the previous hidden state into the corresponding outputs, and $b$ denotes the bias vectors.

Meanwhile, the recurrent unit for GRU at each time step $j$ consists of two gates, i.e., the update gate $z_j$ and the reset gate $r_j$. At each time step $j$, given the time step input $\chi_j$, the hidden state $\eta_j$ is formulated as:
\begin{align}
\eta_j &= {GRU}(\chi_j,\eta_{j-1}) \nonumber \\
 &= (1-z_j) \circ {\eta}_{j-1} + z_j \circ {\underline{\eta}}_j \label{eq:gru}
\end{align}
where
\begin{align*}
{\underline{\eta}}_j &= \tanh(W\chi_j+U[r_j \circ \eta_{j-1}] + b) \\
z_j &= \sigma(W_z\chi_j + U_z\eta_{j-1} + b_z) \\
r_j &= \sigma(W_r\chi_j + U_r\eta_{j-1} + b_r)
\end{align*}
$W$ and $U$ denote the weight matrices transforming the input embedding and the previous hidden state to the corresponding outputs (denoted by the subscript), and $b$ denotes the bias vectors.

As shown in Equations \ref{eq:lstm} and \ref{eq:gru}, both LSTM and GRU make use of gate mechanisms to control the information flow from the input and the hidden state. But unlike LSTM, GRU does not have the memory cell and the output gate. GRU proposes the hidden state $\underline{\eta}_j$ and interpolates each dimension with that of the previous hidden state, controlled by the update gate $z_j$. Meanwhile, the reset gate $r_j$ controls the intensity of the previous hidden state to be taken into account in the current pre-computed hidden state.

The LSTM encoder re-defines Equations \ref{eq:nmtm_encode_fw} and \ref{eq:nmtm_encode_bw} respectively as:
\begin{align*}
\overrightarrow{h}_{j} &= \overrightarrow{g}_{x}(E[x_j],\overrightarrow{h}_{j-1}) = \overrightarrow{LSTM}(E[x_j],\overrightarrow{h}_{j-1}) \\
\overleftarrow{h}_{j} &= \overleftarrow{g}_{x}(E[x_j],\overleftarrow{h}_{j+1}) = \overleftarrow{LSTM}(E[x_j],\overleftarrow{h}_{j+1})
\end{align*}
while the GRU encoder re-defines the two equations respectively as:
\begin{align*}
\overrightarrow{h}_j &= \overrightarrow{g}_{x}(E[x_j],\overrightarrow{h}_{j-1}) = \overrightarrow{GRU}(E[x_j],\overrightarrow{h}_{j-1}) \\
\overleftarrow{h}_j &= \overleftarrow{g}_{x}(E[x_j],\overleftarrow{h}_{j+1}) = \overleftarrow{GRU}(E[x_j],\overleftarrow{h}_{j+1})
\end{align*}
with the embedding representation $E[x_j]$ of word $x_j$ at position $j$ as the input to the recurrent unit function.

For decoding, in Equations \ref{eq:dec_basic} and \ref{eq:dec_basic_att}, both $g_{y,1}$ and $g_{y,2}$ can be instantiated by LSTM (Equation \ref{eq:lstm}) as:
\begin{align*}
g_{y,1}(E[y_{i-1}], s_{i-1}) &= LSTM_1(E[y_{i-1}],s_{i-1}) \\
g_{y,2}(c_{i}, s'_{i}) &= LSTM_2(c_{i},s'_{i})
\end{align*}
or with GRU (Equation \ref{eq:gru}) as:
\begin{align*}
g_{y,1}(E[y_{i-1}], s_{i-1}) &= GRU_1(E[y_{i-1}],s_{i-1}) \\
g_{y,2}(c_{i}, s'_{i}) &= GRU_2(c_{i},s'_{i})
\end{align*}
where the input to the LSTM or GRU function for $g_{y,1}$ is the embedding of the previous word $E[y_{i-1}]$ and the input for  $g_{y,2}$ is the context vector $c_i$.

\subsection{Deep Recurrent Layers}
\label{sec:nmt_deep}

Following \cite{sennrich_university_2017}, we adopt deep RNN models for encoding and decoding. There are two alternatives to achieve this, namely the deep stacked RNN and deep transition RNN. The deep stacked RNN passes the whole input sequence to the first layer of RNN and feeds the sequence of the output hidden representations to the next layer of RNN. This is done subsequently depending on the number of RNN layers. Meanwhile, the deep transition RNN passes an input at each time step through a series of transitions (i.e., recurrent unit functions) and passes the hidden state of the last transition to the next time step.

\subsubsection{Deep Stacked RNN}
\label{sec:nmt_deep_stacked}

\begin{figure*}[htbp]
\centering{
\includegraphics[width=0.8\textwidth]{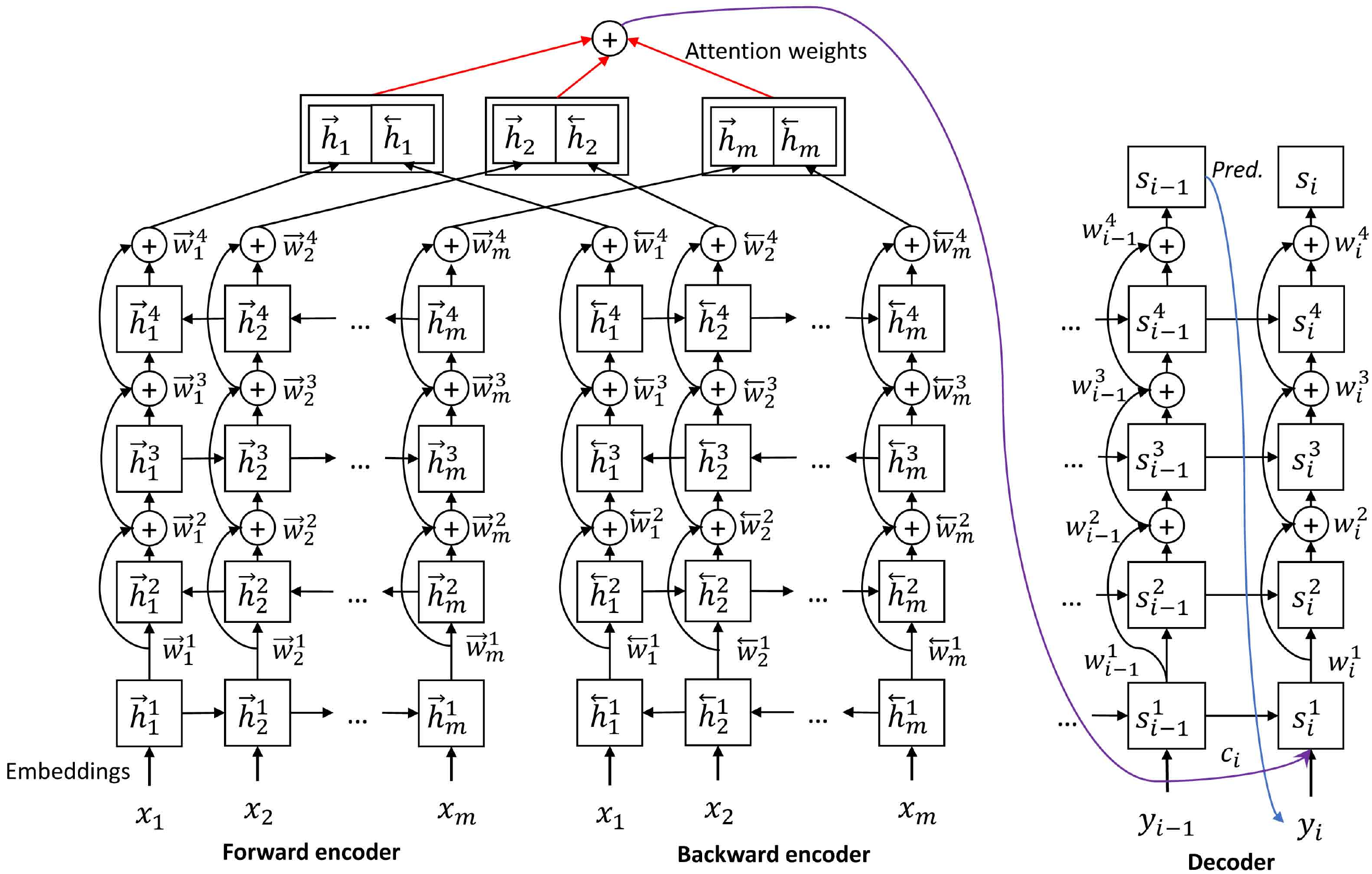}
}
\caption{\label{fig:dstack} An illustration of a deep stacked RNN model \protect\cite{zhou_deep_2016} with encoder stack depth ($D_x$) of 4 and decoder stack depth ($D_y$) of 4.}
\end{figure*}

We adopt the deep stacked RNN that involves residual connection, summing the output of the previous RNN layer with the computed hidden state of the current RNN layer, and alternation of direction \cite{zhou_deep_2016}, as illustrated in Figure \ref{fig:dstack}. In this model, for each layer $l$ and time step $j$, we need to distinguish between the computed hidden state of the current RNN unit function without residual connection, i.e., $\overrightarrow{h}_j^l$, and the hidden state which includes the residual connection, i.e., $\overrightarrow{w}_j^l$. The alternation of direction is designed such that the odd-numbered and even-numbered RNN layers process the sequence in the left-to-right and right-to-left directions respectively. In the stacked RNN with layer depth $D_{x}$, the forward encoder hidden state at time step $j$ in Equation \ref{eq:nmtm_encode_fw}, $\overrightarrow{h}_j$, is computed as follows:
\begin{align*}
\overrightarrow{h}_j &= \overrightarrow{g}_x(E[x_j],\overrightarrow{h}_{j-1}) = \overrightarrow{w}_j^{D_{x}} \\
\overrightarrow{w}_j^{D_{x}} &= \overrightarrow{h}_j^{D_{x}} + \overrightarrow{w}_j^{D_{x}-1} \\
\overrightarrow{w}_j^1 &= \overrightarrow{h}_j^1 = \overrightarrow{g}_x^1(E[x_j],\overrightarrow{h}_{j-1}^1) \\
\overrightarrow{h}_j^{2k} &= \overrightarrow{g}_x^{2k}(\overrightarrow{w}_j^{2k-1}, \overrightarrow{h}_{j+1}^{2k}) \text{, for } 1 < 2k \le D_{x} \\
\overrightarrow{h}_j^{2k+1} &= \overrightarrow{g}_x^{2k+1}(\overrightarrow{w}_j^{2k},\overrightarrow{h}_{j-1}^{2k+1}) \text{, for } 1 < 2k+1 \le D_{x} \\
\overrightarrow{w}_j^l &= \overrightarrow{h}_j^l + \overrightarrow{w}_j^{l-1} \text{, for } 1 < l \le D_{x}
\end{align*}

In the above equations, $\overrightarrow{g}_x^l$ is the forward encoder recurrent unit function of a layer $l$ in the deep RNN stack. It can be instantiated with LSTM or GRU. While the above equations compute the forward encoder hidden state, the backward encoder hidden state $\overleftarrow{h}_j$ is computed similarly by changing the arrow direction from right ($\rightarrow$) to left ($\leftarrow$) and swapping $j-1$ with $j+1$.

Since at each time step, the decoder has no knowledge of the next word, there is no alternation of direction therein. In addition, we only use the recurrent function with attention in the first layer of the RNN stack $s_i^1$, while the deeper layers are simple RNN without attention. Therefore, the decoder RNN with layer depth $D_{y}$ computes the hidden state ${s}_i$ in Equation \ref{eq:nmtm_decode}, as follows:
\begin{align*}
s_i &= g_y(E[y_{i-1}],s_{i-1},c_i) = w_i^{D_{y}} \\
w_i^{D_{y}} &= s_i^{D_{y}} + w_{i}^{D_{y}-1} \\
s_i^{D_{y}} &= g_y^{D_{y}}(w_i^{D_{y}-1},s_{i-1}^{D_{y}}) \\
w_i^1 &= s_i^1 = g_{att}^1(E[y_{i-1}],s_{i-1}^1,c_i) \\
s_i^l &= g_y^l(w_i^{l-1},s_{i-1}^{l}) \text{, for } 1 < l \le D_{y} \\
w_i^l &= s_i^l + w_i^{l-1} \text{, for } 1 < l \le D_{y}
\end{align*}

The first RNN stack layer, $s_i^1 = g_{att}^1(E[y_{i-1}],s_{i-1}^1,c_i)$, is a composition of two recurrent unit functions like in Equations \ref{eq:dec_basic} and \ref{eq:dec_basic_att}, i.e.,
\begin{align*}
s_i^1 &= g_{att}^1(E[y_{i-1}],s_{i-1}^1,c_i) \\
s'_i &= g_{y,1}^1(E[y_{i-1}],s_{i-1}^1) \\
s_i^1 &= g_{y,2}^1(c_i, s'_i)
\end{align*}

Like the encoders, the decoder recurrent unit function $g_y^l$ at each layer $l$ can be instantiated by LSTM or GRU.

\subsubsection{Deep Transition RNN}
\label{sec:nmt_deep_trans}

\begin{figure*}[htbp]
\centering{
\includegraphics[width=0.8\textwidth]{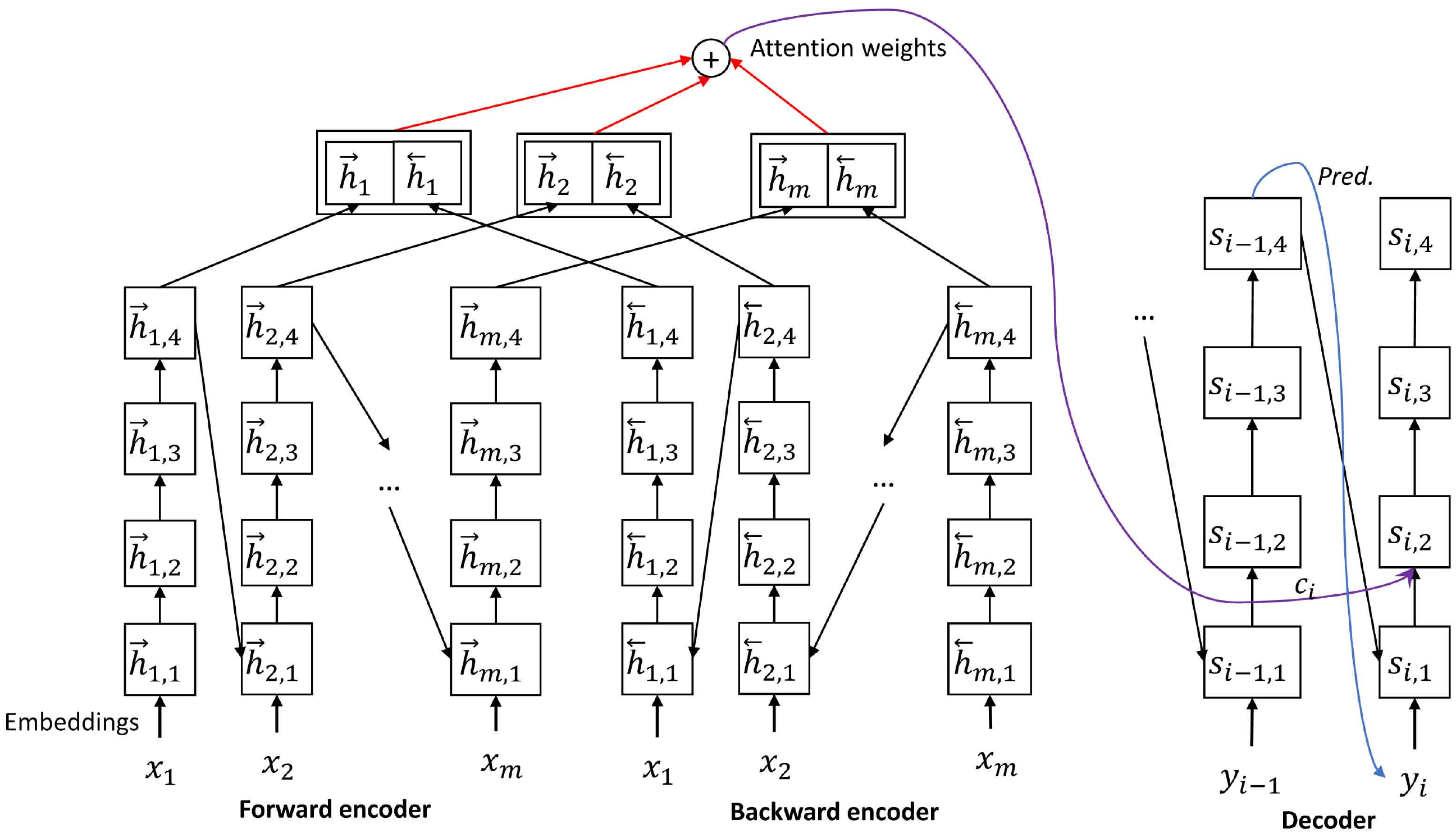}
}
\caption{\label{fig:dtrans} An illustration of a deep transition RNN model \protect\cite{miceli-barone_deep_2017} with 4 encoder transitions ($L_x=4$) and 4 decoder transitions ($L_y=4$).}
\end{figure*}

The deep transition RNN \cite{miceli-barone_deep_2017} involves a number of layers within a time step $j$ through which an input word is fed, as illustrated in Figure \ref{fig:dtrans}. The recurrent unit function of each layer $l$ is defined as a transition, which outputs an intermediate state $\overrightarrow{h}_{j,l}$ for the encoder and $s_{j,l}$ for the decoder. With $L_{x}$ transitions, the hidden state representation at $j$ is equivalent to the output of the last layer, so for the forward encoder state, Equation \ref{eq:nmtm_encode_fw} defines $\overrightarrow{h}_j$ as:
\begin{align*}
\overrightarrow{h}_j &= \overrightarrow{g}_x(E[x_j],\overrightarrow{h}_{j-1}) = \overrightarrow{h}_{j,L_{x}} \\
\overrightarrow{h}_{j,1} &= \overrightarrow{g}_{x,1}(E[x_j], \overrightarrow{h}_{j-1,L_{x}}) \\
\overrightarrow{h}_{j,l} &= \overrightarrow{g}_{x,l}(0, \overrightarrow{h}_{j,l-1}) \text{, for } 1 < l \le L_{x}
\end{align*}
where the input to the first recurrent unit transition is $E[x_j]$, the embedding of the input word at $j$, while the subsequent higher level transitions only receive the output from the previous transition and do not receive the input word. The reverse hidden state $\overleftarrow{h}_j$ in Equation \ref{eq:nmtm_encode_bw} is computed similarly by substituting $j-1$ with $j+1$. The encoder recurrent unit function $\overrightarrow{g}_{x,l}$ at each layer $l$ can be instantiated by LSTM or GRU.

While the baseline shallow decoder RNN already contains two transitions, without and with attention respectively, the deep transition decoder RNN is extended similarly to the deep transition encoder RNN, such that the decoder hidden state with depth $L_{y}$ is computed as:
\begin{align*}
s_{i,1} &= s'_i = {g}_{y,1}(E[y_{i-1}],s_{i-1,L_{y}}) \\
s_{i,2} &= {g}_{y,2}(c_i, s_{i,1}) \\
s_{i,l} &= {g}_{y,l}(0, s_{i,l-1}) \text{, for } 2 < l \le L_{y} \\
s_i &= s_{i,L_{y}}
\end{align*}
Similarly, the decoder recurrent unit function $g_{y,l}$ at each layer $l$ can be instantiated by LSTM or GRU.

\section{Experimental Setup}
\label{sec:setup}

\subsection{Datasets}
\label{sec:setup_data}

We conducted experiments using the parallel training corpora from LDC to test on the NIST test sets. In addition, we also conducted experiments on the United Nations Parallel Corpus \cite{ziemski_united_2016}, following \cite{junczys-dowmunt_is_2016}. We used the pre-defined training, development, and test sets of the corpus following \cite{junczys-dowmunt_is_2016} and conducted NMT experiments accordingly.

We pre-processed our parallel training corpora by segmenting Chinese sentences, which originally have no spaces to demarcate words, and tokenizing English sentences to split punctuation symbols from words. Chinese word segmentation was performed by a maximum entropy model \cite{low_maximum_2005} trained on the Chinese Penn Treebank (CTB) segmentation standard.

To alleviate the effect of rare words in NMT, we fragmented words to sub-words through the byte pair encoding (BPE) algorithm \cite{sennrich_neural_2016} with 59,500 merge operations. All our training sentences are lowercased.

\subsubsection{LDC Corpora}
\label{sec:setup_data_ldc}

We divide the LDC corpora we used into older corpora\footnote{LDC2002E18, LDC2003E14, LDC2004E12, LDC2004T08, LDC2005T06, and LDC2005T10.} and newer corpora\footnote{LDC2007T23, LDC2008T06, LDC2008T08, LDC2008T18, LDC2009T02, LDC2009T06, LDC2009T15, LDC2010T03, LDC2013T11, LDC2013T16, LDC2014T04, LDC2014T11, LDC2014T15, LDC2014T20, and LDC2014T26.}. Due to the dominant older corpora, we duplicate the newer corpora of various domains ten times to achieve better domain balance.

In addition to the parallel sentences, we also utilized a large amount of monolingual English texts, consisting of the English side of FBIS parallel corpus (LDC2003E14) and all the sub-corpora of the English Gigaword Fourth Edition (LDC2009T07). Altogether, the combined corpus consists of 107M sentences and 3.8B tokens. Each individual Gigaword sub-corpus (i.e., AFP, APW, CNA, LTW, NYT, and Xinhua) is used to train a separate $N$-gram language model. The English side of FBIS is also used to train another separate language model (LM). These individual language models are then interpolated to build one single large LM, via perplexity tuning on the English side of the development data. We use this LM for translation output re-ranking.

To recover the original casing on the translation output, we trained a statistical MT recaser model by using Moses \cite{koehn_moses:_2007} on the English side of FBIS parallel text and the Xinhua portion of English Gigaword Fourth Edition.

Due to computation time and memory consideration, parallel sentences in the corpora that are longer than 50 sub-words in either Chinese or English are discarded. In the end, the final parallel training examples consist of 7.65M sentence pairs, 169M Chinese sub-word tokens (equivalent to 166M word tokens), and 186M English sub-word tokens (equivalent to 184M word tokens).

Our translation development (tuning) set is MTC corpus version 1 (LDC2002T01) and version 3 (LDC2004T07). This development set has 1,928 sentence pairs in total, 49K Chinese word tokens and 58K English word tokens on average across the four reference translations. Our translation test set consists of the NIST MT evaluation sets from 2002 to 2006, and 2008\footnote{LDC2010T10, LDC2010T11, LDC2010T12, LDC2010T14, LDC2010T17, and LDC2010T21.}. Altogether in the test sets, there are 7,497 sentence pairs, 192K Chinese word tokens, and 237K English word tokens on average across the four reference translations.

\subsubsection{UN Parallel Corpus}
\label{sec:setup_data_unpc}

The training set of the UN Parallel Corpus, after pre-processing and filtering those exceeding 50 sub-words, consists of 9.73M parallel sentence pairs, 207M Chinese sub-word tokens (equivalent to 204M word tokens), and 225 English sub-word tokens (equivalent to 223M word tokens). We also utilized larger English monolingual text, i.e., all the English side of the UN Parallel Corpus before length filtering, consisting of 11.3M sentences and 335M word tokens. The development set of the UN Parallel Corpus contains 4,000 sentence pairs, 107K Chinese word tokens, and 118K English word tokens, while the test set contains 4,000 sentence pairs, 106K Chinese word tokens, and 118K English word tokens. There is only one reference translation in the development and test sets of the UN Parallel Corpus.

\subsection{NMT Model Parameters}
\label{sec:setup_nmt}

We built our neural machine translation (NMT) system by using Nematus \cite{sennrich_nematus:_2017}, an open-source NMT toolkit which implements the encoder-decoder NMT architecture with attention mechanism. Our system is based on the NMT system in \cite{sennrich_university_2017}. We built an ensemble model consisting of 4 independent models, which are the cross product of two different deep RNN architectures, i.e., deep stacked RNN and deep transition RNN, and two different recurrent unit functions, i.e., GRU and LSTM.

For all our models, the word embedding dimension is 500, and the hidden layer dimension is 1,024. Our deep stacked RNN contains 4 stack layers on each of the encoder and decoder. Meanwhile, our deep transition RNN contains 4 encoder transitions and 8 decoder transitions.

Training for each individual model progresses by updating the model parameters at each mini-batch of 40 sentence pairs to minimize the negative log-likelihood loss function on the parallel training data. We use the Adam algorithm \cite{kingma_adam:_2015} with learning rate of $0.0001$. At each update, we clip the gradient norm to $1.0$. We apply layer normalization \cite{ba_layer_2016} on the model parameters for faster convergence and tie the target-side embedding with the transpose of the output weight matrix \cite{press_using_2017}. Model parameters are saved at every checkpoint of 10,000 update iterations. At this stage, the negative log-likelihood loss function on the development set is checked. Training stops when there has been no improvement over the lowest loss function value on the development set for 10 consecutive checkpoints.

The main difference between our system and \cite{sennrich_university_2017} is that while they only built NMT models with GRU, we also made use of LSTM. Another difference is in the usage of the larger monolingual English text. They built a synthetic Chinese-English parallel corpus by translating the monolingual English text to Chinese with an English-to-Chinese (reverse direction) NMT model and appended those sentence pairs to their parallel training corpus. Meanwhile, we exploited the English monolingual corpus by building a 5-gram language model to re-rank the $k$-best translation outputs produced by our NMT system.

\section{Experimental Results}
\label{sec:results}

\begin{table*}[htbp]
\centering
\begin{tabular}{|c||l|l|l|l||l|l|}
\hline
\multirow{2}{*}{\textbf{Dataset}} & \multicolumn{2}{c|}{\textbf{GRU}} & \multicolumn{2}{c||}{\textbf{LSTM}} & \multicolumn{2}{c|}{\textbf{4mod-ens}} \\\cline{2-7}
 & \multicolumn{1}{c|}{\textbf{dstack}} & \multicolumn{1}{c|}{\textbf{dtrans}} & \multicolumn{1}{c|}{\textbf{dstack}} & \multicolumn{1}{c||}{\textbf{dtrans}} & \multicolumn{1}{c|}{\textbf{no re-ranking}} & \multicolumn{1}{c|}{\textbf{with re-ranking}}  \\\hline\hline
NIST02 & 41.27 & 43.28 & 44.03 & 42.12 & 46.82$^{**}$ & 46.94  \\\hline
NIST03 & 41.78 & 42.38 & 42.49 & 41.87 & 47.42$^{**}$ & 47.58  \\\hline
NIST04 & 43.75 & 44.33 & 45.11 & 43.97 & 49.12$^{**}$ & 49.13  \\\hline
NIST05 & 41.71 & 42.52 & 43.40 & 41.97 & 47.72$^{**}$ & 47.78  \\\hline
NIST06 & 42.27 & 43.18 & 43.43 & 42.19 & 49.19$^{**}$ & 49.37  \\\hline
NIST08 & 35.28 & 36.11 & 36.78 & 35.55 & 41.36$^{**}$ & 41.48  \\\hline
\hline
Average & 41.01 & 41.97 & \textbf{42.54} & 41.28 & \textbf{46.94}$^{**}$ & \textbf{47.05}  \\
\hline
\end{tabular}
\caption{\label{tab:main-result} Experimental results in BLEU (\%) of our NMT systems on NIST data set from LDC. Each individual model is obtained by cross-combining two different deep RNN architectures, i.e., deep stacked (\textbf{dstack}) and deep transition (\textbf{dtrans}) RNN, with two different recurrent unit functions, i.e., \textbf{GRU} and \textbf{LSTM}, without $k$-best re-ranking. The ensemble of 4 model types (\textbf{4mod-ens}) is obtained by taking the best model from each individual model type. This setting is tested both without and with re-ranking. Statistical significance testing was done to compare \textbf{4mod-ens} with the best individual model type, \textbf{dstack-LSTM} ($**$: significant at $p<0.01$).}
\end{table*}

For experiments using the LDC corpora, translation quality is measured by case-insensitive BLEU \cite{papineni_bleu:_2002}, for which the brevity penalty is computed based on the shortest reference (NIST-BLEU)\footnote{\url{ftp://jaguar.ncsl.nist.gov/mt/resources/mteval-v11b.pl}}. Statistical significance testing between systems is conducted by bootstrap resampling \cite{koehn_statistical_2004}.

As shown in Table \ref{tab:main-result}, among the individual model types, the deep stacked LSTM NMT model gives the best performance. However the best result is achieved by the ensemble of 4 independent model types, combining deep stacked and deep transition architectures with GRU and LSTM recurrent unit functions. This ensemble model gives an improvement of 4.40 BLEU points over the best deep stacked LSTM model. Applying re-ranking by the $N$-gram language model on top of the ensemble system of 4 independent models gives a further improvement of 0.11 BLEU point on average, which gives the best result of our system.

% 4-vs-2 comparison
\begin{table}[htbp]
\centering
\begin{tabular}{|c|l|l||l|}
\hline
\multirow{2}{*}{\backslashbox{\bf Unit}{\bf Layer}} & \multicolumn{1}{c|}{\multirow{2}{*}{\bf dstack}} & \multicolumn{1}{c||}{\multirow{2}{*}{\bf dtrans}} & \multicolumn{1}{c|}{\bf dstack} \\
 & & & \multicolumn{1}{c|}{\bf +dtrans} \\\hline\hline
{\bf GRU} & 41.01 & 41.97 & 45.15 \\\hline
{\bf LSTM} & 42.54 & 41.28 & 45.34 \\\hline\hline
{\bf GRU+LSTM} & 44.72 & 45.02 & {\bf 46.94}$^{**\dagger\dagger\ddagger\ddagger\#\#}$ \\\hline
\end{tabular}
\caption{\label{tab:abl-result} Experimental results in BLEU (\%) on NIST data set showing different combinations of RNN {\bf unit} functions and  deep {\bf layers}. The RNN unit functions include GRU, LSTM, and the ensemble of the two. The deep layers include deep stacked ({\bf dstack}) and deep transition ({\bf dtrans}) RNN, and the ensemble of the two. Statistical significance testing is shown to compare our 4-model ensemble with the ensemble of {\bf dstack+dtrans GRU} ($**$: significant at $p<0.01$), with {\bf dstack+dtrans LSTM} ($\dagger\dagger$: significant at $p<0.01$), with {\bf dstack GRU+LSTM} ($\ddagger\ddagger$: significant at $p<0.01$), and with {\bf dtrans GRU+LSTM} ($\#\#$: significant at $p<0.01$).}
\end{table}

We are interested in comparing our 4-model ensemble before re-ranking with the ensemble of 2 models with only one RNN unit function (but two different deep layers) and with only one deep layer (but two different RNN units). As shown in Table \ref{tab:abl-result}, our 4-model ensemble outperforms every 2-model ensemble, and the improvement is statistically significant ($p < 0.01$).

We also compare the best results of our system on NIST test sets with the best results reported in published papers on Chinese-to-English MT systems in major computational linguistics and artificial intelligence publication venues\footnote{CL journal, TACL, ACL, COLING, EMNLP, NAACL, SSST, WMT, AAAI, and IJCAI.}. Over the years 2007--2017 (both years inclusive), our 4-model ensemble system with re-ranking achieves a higher BLEU score than the best results reported in almost all (402 out of 403) papers\footnote{In one paper out of 403 papers \protect\cite{huang_factored_2013}, testing was performed on 4 test subsets: the news subset and the web subset of NIST06 and NIST08. Our system is only marginally worse (by less than 0.1 BLEU point) on the news subset of NIST06, and is better by a large margin (by 2--6 BLEU points) on the other 3 subsets.}.

Note that the LDC training datasets used in the published papers that we compare to may not be the same as ours. This incomparability would not have happened if there were a widely adopted, standardized dataset for training.

In addition, since there are two ways of computing BLEU scores with respect to word casing, i.e., case-insensitive and case-sensitive, we have taken care to compare BLEU scores using the same word casing, that is, by comparing our case-sensitive BLEU scores only to case-sensitive BLEU scores published previously, and similarly for case-insensitive BLEU scores.

Moreover, there are two ways of computation with respect to brevity penalty calculation involving multiple reference translations, namely ``shortest'', used in NIST-BLEU, and ``closest'', used in IBM-BLEU. The former sets brevity penalty against the shortest reference translation while the latter sets brevity penalty against the reference translation whose length is the most similar to the system translation output. We have also taken this into account by comparing NIST-BLEU with NIST-BLEU, and IBM-BLEU with IBM-BLEU\footnote{\url{ftp://jaguar.ncsl.nist.gov/mt/resources/mteval-v13a.pl}}.

For UN Parallel Corpus experiments, translation quality is measured by case-insensitive BLEU using the script provided by Moses\footnote{The \url{multi-bleu.perl} script in the Moses distribution.}, following the evaluation setup in \cite{junczys-dowmunt_is_2016}\footnote{Personal communication.}.

\begin{table}[htbp]
\centering{
\begin{tabular}{|l|l|l|}
\hline
\multicolumn{1}{|c|}{\multirow{2}{*}{\bf Published}} & \multicolumn{2}{c|}{\bf Ours (4mod-ens)} \\\cline{2-3}
 & {\bf no reranking} & {\bf with reranking} \\\hline
53.1 & 55.0 & 55.3$^{*}$ \\\hline
\end{tabular}
}
\caption{\label{tab:unpc-result} Experimental results in BLEU (\%) on the test set of the UN Parallel Corpus of the best published result in \protect\cite{junczys-dowmunt_is_2016} and our system, without and with $k$-best re-ranking with $N$-gram language model. Statistical significance testing shows the comparison between our 4-model system with re-ranking and without re-ranking ($*$: significant at $p<0.05$).}
\end{table}

As shown in Table \ref{tab:unpc-result}, our best result is obtained by an ensemble of 4 independent models with $k$-best output re-ranking using $N$-gram LM trained on the whole English side of the UN Parallel Corpus. Our system with re-ranking is 0.3 BLEU point better than without re-ranking, and the improvement is statistically significant ($p < 0.05$). Both of our systems achieve higher BLEU scores than the best published result for Chinese-to-English translation reported in \cite{junczys-dowmunt_is_2016}.

\section{Related Work}
\label{sec:related}

Establishing standards for the state of the art by publicly accessible resources is important in research. In speech recognition, for instance, there has been work on building a virtual machine as a means of collaboratively building a state-of-the-art system for speech recognition \cite{metze_speech_2013}, aiming at realizing a standardized state-of-the-art system in a collaborative manner. While we do not provide any virtual machines, we have a similar intention of making available a state-of-the-art MT system.

On NMT, \newcite{denkowski_stronger_2017} argued that experiments should be based on a strong baseline system to ensure that a newly proposed approach indeed improves over the best prior published approaches. They performed their experiments on WMT and IWSLT tasks (but not on Chinese-to-English translation) which have fixed training, development, and test sets. The problem for Chinese-to-English MT is greater in that there is no pre-defined set of training data that must be used for experiments, and various groups used different tuning sets and reported their results on different NIST OpenMT test sets. In addition, the lack of a standardized benchmark is not limited to neural MT approaches, but has been widespread since statistical MT approaches began to be tested on Chinese-to-English translation.

\section{Conclusion}
\label{sec:conclusion}

The problem of lack of consistent experimental setups for Chinese-to-English MT poses a challenge in evaluating newly proposed approaches over the pre-existing state of the art. This can be avoided if there is a clear benchmark which consists of standardized training, development, and test sets, as well as a common benchmark system setup. In this paper, we have shown that our proposed MT approach can be used to build a competitive system on the NIST OpenMT test sets that outperforms systems in almost all 403 published papers in the past 11 years.

We encourage Chinese-to-English MT experiments to use our common benchmark consisting of standard data and evaluation. As the NIST dataset from LDC is widely used in the Chinese-to-English MT research community, we have put up a scoreboard listing the scores achieved by all prior published papers when evaluated on the NIST dataset and released the source code and translation output of our best NMT system\footnote{\url{https://github.com/nusnlp/c2e-mt-benchmark}}. By doing so, we hope future work can make more meaningful comparisons to previous MT research.

\vspace{.3\baselineskip}

% \nocite{*}
\section{Bibliographical References}
\label{main:ref}

\bibliographystyle{lrec}
\bibliography{uta}

\begin{thebibliography}{}

\bibitem[\protect\citename{Ba \bgroup et al.\egroup }2016]{ba_layer_2016}
Ba, J.~L., Kiros, J.~R., and Hinton, G.~E.
\newblock (2016).
\newblock Layer normalization.
\newblock {\em CoRR}, abs/1607.06450.

\bibitem[\protect\citename{Bahdanau \bgroup et al.\egroup
  }2015]{bahdanau_neural_2015}
Bahdanau, D., Cho, K., and Bengio, Y.
\newblock (2015).
\newblock Neural machine translation by jointly learning to align and
  translate.
\newblock In {\em Proceedings of the 3rd {International} {Conference} on
  {Learning} {Representations} ({ICLR} 2015)}.

\bibitem[\protect\citename{Cho \bgroup et al.\egroup }2014]{cho_learning_2014}
Cho, K., van Merri{\"e}nboer, B., Gulcehre, C., Bahdanau, D., Bougares, F.,
  Schwenk, H., and Bengio, Y.
\newblock (2014).
\newblock Learning phrase representations using {RNN} encoder-decoder for
  statistical machine translation.
\newblock In {\em Proceedings of the 2014 {Conference} on {Empirical} {Methods}
  in {Natural} {Language} {Processing}}, pages 1724--1734.

\bibitem[\protect\citename{Denkowski and Neubig}2017]{denkowski_stronger_2017}
Denkowski, M. and Neubig, G.
\newblock (2017).
\newblock Stronger baselines for trustable results in neural machine
  translation.
\newblock In {\em Proceedings of the {First} {Workshop} on {Neural} {Machine}
  {Translation}}, pages 18--27.

\bibitem[\protect\citename{Hochreiter and
  Schmidhuber}1997]{hochreiter_long_1997}
Hochreiter, S. and Schmidhuber, J.
\newblock (1997).
\newblock Long short-term memory.
\newblock {\em Neural Computation}, 9(8):1735--1780.

\bibitem[\protect\citename{Huang \bgroup et al.\egroup
  }2013]{huang_factored_2013}
Huang, Z., Devlin, J., and Zbib, R.
\newblock (2013).
\newblock Factored soft source syntactic constraints for hierarchical machine
  translation.
\newblock In {\em Proceedings of the 2013 {Conference} on {Empirical} {Methods}
  in {Natural} {Language} {Processing}}, pages 556--566.

\bibitem[\protect\citename{Junczys-Dowmunt \bgroup et al.\egroup
  }2016]{junczys-dowmunt_is_2016}
Junczys-Dowmunt, M., Dwojak, T., and Hoang, H.
\newblock (2016).
\newblock Is neural machine translation ready for deployment? {A} case study on
  30 translation directions.
\newblock In {\em Proceedings of the 13th {International} {Workshop} on
  {Spoken} {Language} {Translation}}.

\bibitem[\protect\citename{Kingma and Ba}2015]{kingma_adam:_2015}
Kingma, D.~P. and Ba, J.~L.
\newblock (2015).
\newblock Adam: {A} method for stochastic optimization.
\newblock In {\em Proceedings of the 3rd {International} {Conference} on
  {Learning} {Representations} ({ICLR} 2015)}.

\bibitem[\protect\citename{Koehn \bgroup et al.\egroup
  }2007]{koehn_moses:_2007}
Koehn, P., Hoang, H., Birch, A., Callison-Burch, C., Federico, M., Bertoldi,
  N., Cowan, B., Shen, W., Moran, C., Zens, R., Dyer, C., Bojar, O.,
  Constantin, A., and Herbst, E.
\newblock (2007).
\newblock Moses: {Open} source toolkit for statistical machine translation.
\newblock In {\em Proceedings of the {ACL} 2007 {Demo} and {Poster}
  {Sessions}}, pages 177--180.

\bibitem[\protect\citename{Koehn}2004]{koehn_statistical_2004}
Koehn, P.
\newblock (2004).
\newblock Statistical significance tests for machine translation evaluation.
\newblock In {\em Proceedings of the 2004 {Conference} on {Empirical} {Methods}
  in {Natural} {Language} {Processing}}, pages 388--395.

\bibitem[\protect\citename{Low \bgroup et al.\egroup }2005]{low_maximum_2005}
Low, J.~K., Ng, H.~T., and Guo, W.
\newblock (2005).
\newblock A maximum entropy approach to {Chinese} word segmentation.
\newblock In {\em Proceedings of the {Fourth} {SIGHAN} {Workshop} on {Chinese}
  {Language} {Processing}}, pages 161--164.

\bibitem[\protect\citename{Metze \bgroup et al.\egroup
  }2013]{metze_speech_2013}
Metze, F., Fosler-Lussier, E., and Bates, R.
\newblock (2013).
\newblock The speech recognition virtual kitchen.
\newblock In {\em Proceedings of the 14th {Annual} {Conference} of the
  {International} {Speech} {Communication} {Association} ({Interspeech}-2013)},
  pages 1858--1860.

\bibitem[\protect\citename{Miceli-Barone \bgroup et al.\egroup
  }2017]{miceli-barone_deep_2017}
Miceli-Barone, A.~V., Helcl, J., Sennrich, R., Haddow, B., and Birch, A.
\newblock (2017).
\newblock Deep architectures for neural machine translation.
\newblock In {\em Proceedings of the {Second} {Conference} on {Machine}
  {Translation}}, pages 99--107.

\bibitem[\protect\citename{Papineni \bgroup et al.\egroup
  }2002]{papineni_bleu:_2002}
Papineni, K., Roukos, S., Ward, T., and Zhu, W.-J.
\newblock (2002).
\newblock {BLEU}: {A} method for automatic evaluation of machine translation.
\newblock In {\em Proceedings of the 40th {Annual} {Meeting} of the
  {Association} for {Computational} {Linguistics}}, pages 311--318.

\bibitem[\protect\citename{Press and Wolf}2017]{press_using_2017}
Press, O. and Wolf, L.
\newblock (2017).
\newblock Using the output embedding to improve language models.
\newblock In {\em Proceedings of the 15th {Conference} of the {European}
  {Chapter} of the {Association} for {Computational} {Linguistics}: {Short}
  {Papers}}, pages 157--163.

\bibitem[\protect\citename{Sennrich \bgroup et al.\egroup
  }2016]{sennrich_neural_2016}
Sennrich, R., Haddow, B., and Birch, A.
\newblock (2016).
\newblock Neural machine translation of rare words with subword units.
\newblock In {\em Proceedings of the 54th {Annual} {Meeting} of the
  {Association} for {Computational} {Linguistics}}, pages 1715--1725.

\bibitem[\protect\citename{Sennrich \bgroup et al.\egroup
  }2017a]{sennrich_university_2017}
Sennrich, R., Birch, A., Currey, A., Germann, U., Haddow, B., Heafield, K.,
  Miceli-Barone, A.~V., and Williams, P.
\newblock (2017a).
\newblock The {University} of {Edinburgh}'s neural {MT} systems for {WMT}17.
\newblock In {\em Proceedings of the {Conference} on {Machine} {Translation}:
  {Shared} {Task} {Papers}}, pages 389--399.

\bibitem[\protect\citename{Sennrich \bgroup et al.\egroup
  }2017b]{sennrich_nematus:_2017}
Sennrich, R., Firat, O., Cho, K., Birch, A., Haddow, B., Hitschler, J.,
  Junczys-Dowmunt, M., L{\"a}ubli, S., Miceli-Barone, A.~V., Mokry, J., and
  N{\u a}dejde, M.
\newblock (2017b).
\newblock Nematus: a toolkit for neural machine translation.
\newblock In {\em Proceedings of the {Software} {Demonstrations} of the 15th
  {Conference} of the {European} {Chapter} of the {Association} for
  {Computational} {Linguistics}}, pages 65--68.

\bibitem[\protect\citename{Zhou \bgroup et al.\egroup }2016]{zhou_deep_2016}
Zhou, J., Cao, Y., Wang, X., Li, P., and Xu, W.
\newblock (2016).
\newblock Deep recurrent models with fast-forward connections for neural
  machine translation.
\newblock {\em Transactions of the Association for Computational Linguistics},
  4(1):371--383.

\bibitem[\protect\citename{Ziemski \bgroup et al.\egroup
  }2016]{ziemski_united_2016}
Ziemski, M., Junczys-Dowmunt, M., and Pouliquen, B.
\newblock (2016).
\newblock The {United} {Nations} {Parallel} {Corpus} v1.0.
\newblock In {\em Proceedings of the {Tenth} {International} {Conference} on
  {Language} {Resources} and {Evaluation} ({LREC} 2016)}.

\end{thebibliography}

%\section{Language Resource References}
%\label{lr:ref}
%\bibliographystylelanguageresource{lrec}
%\bibliographylanguageresource{uta}

\end{document}